\documentclass[conference]{IEEEtran}
\IEEEoverridecommandlockouts
\usepackage{cite}
\usepackage{amsmath,amssymb,amsfonts}
\usepackage{algorithmic}
\usepackage{graphicx}
\usepackage{textcomp}
\usepackage{xcolor}

\usepackage{algorithm}
\usepackage{array}
\usepackage{booktabs}
\usepackage{enumitem}
\usepackage{pgfplots}
\usepackage{colortbl}
\definecolor{grey}{rgb}{0.9, 0.9, 0.9}
\usepackage{stmaryrd}
\usepackage{booktabs}
\usepackage{multirow} 
\usepackage[flushleft]{threeparttable}
\usepackage{capt-of}
\usepackage[export]{adjustbox}  
\usepackage[percent]{overpic}  
\usepackage{arydshln}
\usepackage{pifont}

\usepackage{textcomp}
\usepackage{stfloats}
\usepackage{url}
\usepackage{verbatim}
\usepackage{blindtext}
\usepackage{soul}
\usepackage{gensymb}
\usepackage{subcaption}
\usepackage{hhline}
\usepackage{tikz}
\usetikzlibrary{positioning,fit,calc,spy}

\pgfplotsset{compat=1.18}

\usepackage[pagebackref,breaklinks,colorlinks]{hyperref}

\makeatletter
\renewcommand{\paragraph}{%
	\@startsection{paragraph}{4}%
	{\z@}{0.65ex \@plus 1ex \@minus .2ex}{-1em}
	{\normalfont \normalsize \bfseries }%
}
\makeatother
\definecolor{col1}{HTML}{a4a0a6}
\definecolor{col2}{HTML}{e5836f} 
\definecolor{col3}{HTML}{808080}
\newcommand\videozoomed[3]{
    \begin{tikzpicture}[
    zoomboxarray,
    zoomboxes below,
    connect zoomboxes, zoomboxarray columns/.initial=2,
    zoombox paths/.append style={thick}]
        \node[image node]{\includegraphics[width=0.19\textwidth]{images/dynerf/#1}};
        
        \def\ip{#2}
        \def\null{X}
        \ifx\ip\null
            \def\donothing{0}
        \else
            \node[align=left, white] at (0.6, 2.3){#2MB};
            \node[align=left, white] at (0.25, 2.0){#3x};
        \fi

        \zoombox[magnification=5, color code=col1]{0.45,0.19} 
        \zoombox[magnification=6, color code=col2]{0.693,0.113} %
    \end{tikzpicture}
}

\definecolor{col3}{HTML}{a4a0a6}
\definecolor{col4}{HTML}{e5836f}

\newcommand{\plotcroptrex}[5]{%
    \begin{tikzpicture} 
        \node{\adjincludegraphics[trim={{0.4\width} {0.34\height} {0.3\width} {0.25\height}}, clip, width=\linewidth] {images/dnerf/trex/#1}};
        
        \def\ip{#2}
        \def\null{X}
        \def\first{200MB}
        \ifx\ip\null
            \def\donothing{0}
        \else %
            \ifx\ip\first
                \node[align=right, black] at (-.4, 1.6){\scriptsize Size:};
                \node[align=right, black] at (-.1, 1.3){\scriptsize PSNR:};
                \node[align=left, black] at (.1, 1.){\scriptsize SSIM:};
            \else
                \def\donothing{0}
            \fi
            \node[align=right, black] at (.95, 1.6){\scriptsize #2};
            \node[align=right, black] at (0.955, 1.3){\scriptsize #4};
            \node[align=left, black] at (.95, 1.){\scriptsize #5};
        \fi

        \node[align=left, black] at (-0., -2.2){\scriptsize #3};
    \end{tikzpicture}
}

\definecolor{col1}{HTML}{000000}

\newcommand{\plotcrophook}[3]{%
    \begin{tikzpicture}
        \node{\adjincludegraphics[trim={{0.4\width} {0.65\height} {0.3\width} {0.1\height}}, clip, width=\linewidth]{images/dnerf/hook/#1.jpg}};
        \def\ip{#2}
        \def\null{X}
        \ifx\ip\null
            \def\donothing{0};
        \else
            \node[align=left, black] at (-0.8, 1.){\scriptsize #2MB};
        \fi
        \node[align=left, black] at (-0.3, -1.5){\small #3};
    \end{tikzpicture}
    
}

\newcommand{\plotlego}[4]{%
    \begin{tikzpicture}
        \node{\adjincludegraphics[width=\linewidth]{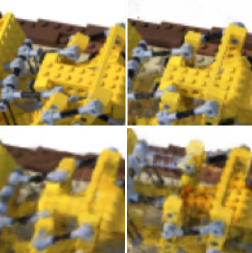}};
        \node[align=left, black] at (-2.2, -.1){\scriptsize Ground Truth}; 
        \node[align=left, black] at (2.2, -.1){\scriptsize Ours-NeRF - 25.31 / 0.94 / 9.7MB}; 
        \node[align=left, black] at (2.2, -4.6){\scriptsize D-NeRF - 21.64 / 0.83 / 13MB};
        \node[align=left, black] at (-2.2, -4.6){\scriptsize TiNeuVox - 24.35 / 0.88 / 8MB}; 
        
    \end{tikzpicture}
    
}
\newcommand{\plotmorezoomed}[1]{%
    \begin{tikzpicture}
        \node{\adjincludegraphics[width=\linewidth]{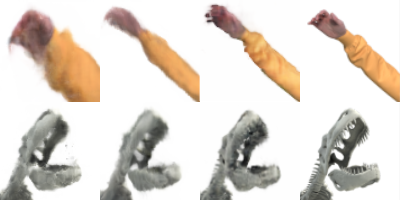}};
        \node[align=left, black] at (-3.2, -2.6){\scriptsize D-NeRF};
        \node[align=left, black] at (-1.1, -2.6){\scriptsize TiNeuVox};
        \node[align=left, black] at (1., -2.6){\scriptsize Ours-NeRF};
        \node[align=left, black] at (3.1, -2.6){\scriptsize Ground Truth};
        
    \end{tikzpicture}
    
}

\newcommand{\plotkp}[1]{%
    \begin{tikzpicture}
        \node{\adjincludegraphics[width=0.98\linewidth]{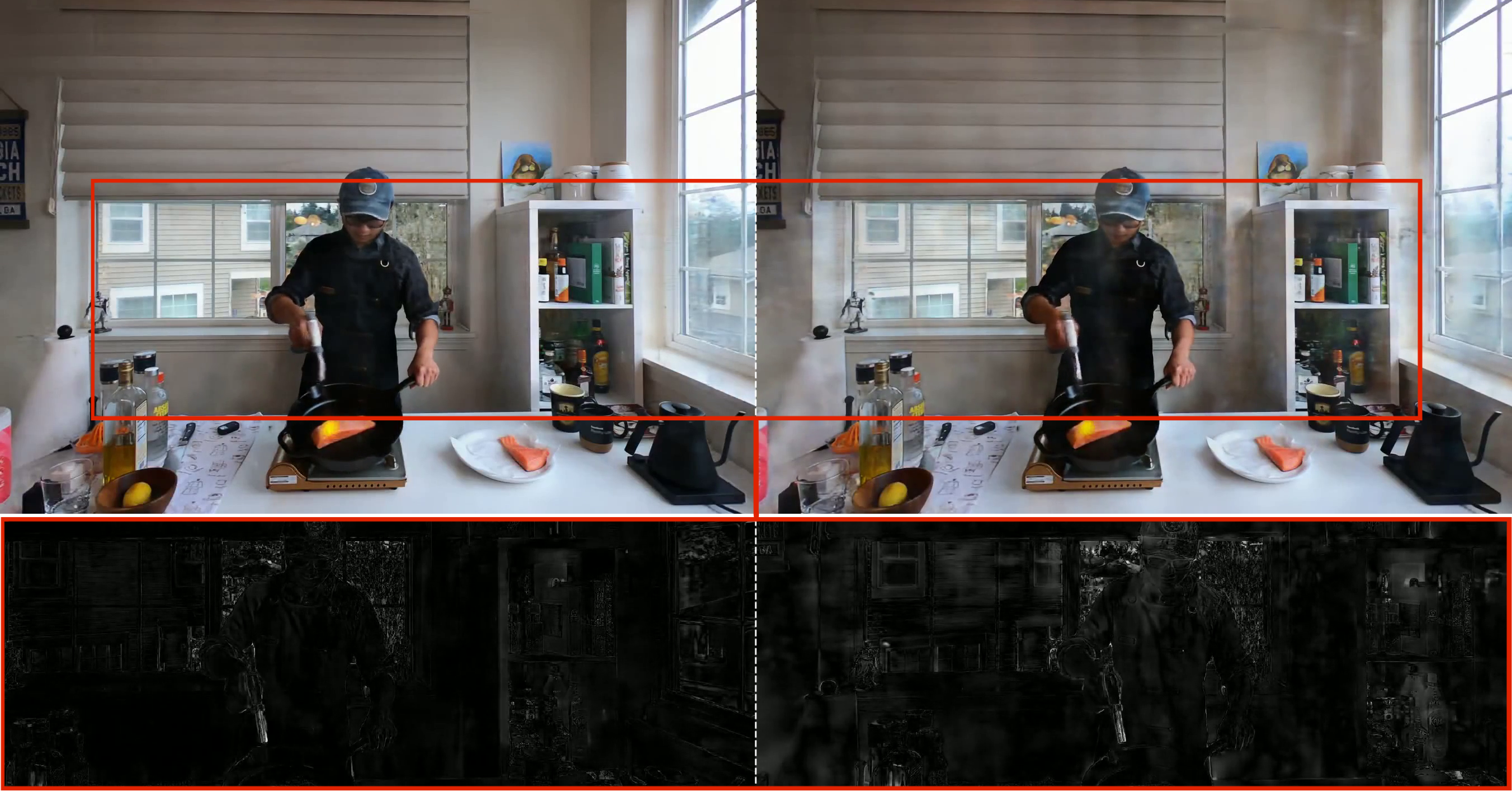}};
        \node[align=left, white] at (2., 2.){\scriptsize PSNR: 25.27 SSIM: 0.871};
        \node[align=left, white] at (-2., 2.){\scriptsize PSNR: 28.25 SSIM: 0.900};

        \node[align=left, black] at (2.3, -2.6){\scriptsize K-Planes-Compact};
        \node[align=left, black] at (-2.3, -2.6){\scriptsize Ours-NeRF};
    \end{tikzpicture}
    
}

\newif\ifblackandwhitecycle
\gdef\patternnumber{0}

\pgfkeys{/tikz/.cd,
    zoombox paths/.style={
        draw=orange,
        very thick
    },
    black and white/.is choice,
    black and white/.default=static,
    black and white/static/.style={
        draw=white,
        zoombox paths/.append style={
            draw=white,
            postaction={
                draw=black,
                loosely dashed
            }
        }
    },
    black and white/static/.code={
        \gdef\patternnumber{1}
    },
    black and white/cycle/.code={
        \blackandwhitecycletrue
        \gdef\patternnumber{1}
    },
    black and white pattern/.is choice,
    black and white pattern/0/.style={},
    black and white pattern/1/.style={
            draw=white,
            postaction={
                draw=black,
                dash pattern=on 2pt off 2pt
            }
    },
    black and white pattern/2/.style={
            draw=white,
            postaction={
                draw=black,
                dash pattern=on 4pt off 4pt
            }
    },
    black and white pattern/3/.style={
            draw=white,
            postaction={
                draw=black,
                dash pattern=on 4pt off 4pt on 1pt off 4pt
            }
    },
    black and white pattern/4/.style={
            draw=white,
            postaction={
                draw=black,
                dash pattern=on 4pt off 2pt on 2 pt off 2pt on 2 pt off 2pt
            }
    },
    zoomboxarray inner gap/.initial=5pt,
    zoomboxarray columns/.initial=2,
    zoomboxarray rows/.initial=1,
    zoomboxarray heightmultiplier/.initial=0.5,
    subfigurename/.initial={},
    figurename/.initial={zoombox},
    zoomboxarray/.style={
        execute at begin picture={
            \begin{scope}[
                spy using outlines={%
                    zoombox paths,
                    width=\imagewidth / \pgfkeysvalueof{/tikz/zoomboxarray columns} - (\pgfkeysvalueof{/tikz/zoomboxarray columns} - 1) / \pgfkeysvalueof{/tikz/zoomboxarray columns} * \pgfkeysvalueof{/tikz/zoomboxarray inner gap} -\pgflinewidth,
                    height=\pgfkeysvalueof{/tikz/zoomboxarray heightmultiplier} * (\imageheight / \pgfkeysvalueof{/tikz/zoomboxarray rows} - (\pgfkeysvalueof{/tikz/zoomboxarray rows} - 1) / \pgfkeysvalueof{/tikz/zoomboxarray rows} * \pgfkeysvalueof{/tikz/zoomboxarray inner gap}-\pgflinewidth),
                    magnification=3,
                    every spy on node/.style={
                        zoombox paths
                    },
                    every spy in node/.style={
                        zoombox paths
                    }
                }
            ]
        },
        execute at end picture={
            \end{scope}
     \gdef\patternnumber{0}
        },
        spymargin/.initial=0.2em,
        zoomboxes xshift/.initial=1,
        zoomboxes right/.code=\pgfkeys{/tikz/zoomboxes xshift=1},
        zoomboxes left/.code=\pgfkeys{/tikz/zoomboxes xshift=-1},
        zoomboxes yshift/.initial=0,
        zoomboxes above/.code={
            \pgfkeys{/tikz/zoomboxes yshift=1},
            \pgfkeys{/tikz/zoomboxes xshift=0}
        },
        zoomboxes below/.code={
            \pgfkeys{/tikz/zoomboxes yshift=-1},
            \pgfkeys{/tikz/zoomboxes xshift=0}
        },
        caption margin/.initial=0ex, %
    },
    adjust caption spacing/.code={},
    image container/.style={
        inner sep=0pt,
        at=(image.north),
        anchor=north,
        adjust caption spacing
    },
    zoomboxes container/.style={
        inner sep=0pt,
        at=(image.north),
        anchor=north,
        name=zoomboxes container,
        xshift=\pgfkeysvalueof{/tikz/zoomboxes xshift}*(\imagewidth+\pgfkeysvalueof{/tikz/spymargin}),
        yshift=\pgfkeysvalueof{/tikz/zoomboxes yshift}*(\imageheight+\pgfkeysvalueof{/tikz/spymargin}+\pgfkeysvalueof{/tikz/caption margin}),
        adjust caption spacing
    },
    calculate dimensions/.code={
        \pgfpointdiff{\pgfpointanchor{image}{south west} }{\pgfpointanchor{image}{north east} }
        \pgfgetlastxy{\imagewidth}{\imageheight}
        \global\let\imagewidth=\imagewidth
        \global\let\imageheight=\imageheight
        \gdef\columncount{1}
        \gdef\rowcount{1}
        
    },
    image node/.style={
        inner sep=0pt,
        name=image,
        anchor=south west,
        append after command={
            [calculate dimensions]
            node [image container,subfigurename=\pgfkeysvalueof{/tikz/figurename}-image] {\phantomimage}
            node [zoomboxes container,subfigurename=\pgfkeysvalueof{/tikz/figurename}-zoom] {\phantomimage}
        }
    },
    color code/.style={
        zoombox paths/.append style={draw=#1}
    },
    connect zoomboxes/.style={
    spy connection path={\draw[draw=none,zoombox paths] (tikzspyonnode) -- (tikzspyinnode);}
    },
    help grid code/.code={
        \begin{scope}[
                x={(image.south east)},
                y={(image.north west)},
                font=\footnotesize,
                help lines,
                overlay
            ]
            \foreach \x in {0,1,...,9} {
                \draw(\x/10,0) -- (\x/10,1);
                \node [anchor=north] at (\x/10,0) {0.\x};
            }
            \foreach \y in {0,1,...,9} {
                \draw(0,\y/10) -- (1,\y/10);                        \node [anchor=east] at (0,\y/10) {0.\y};
            }
        \end{scope}
    },
    help grid/.style={
        append after command={
            [help grid code]
        }
    },
}

\newcommand\phantomimage{%
    \phantom{%
        \rule{\imagewidth}{\imageheight}%
    }%
}
\newcommand\zoombox[2][]{
    \begin{scope}[zoombox paths]
        \pgfmathsetmacro\xpos{
            (\columncount-1)*(\imagewidth / \pgfkeysvalueof{/tikz/zoomboxarray columns} + \pgfkeysvalueof{/tikz/zoomboxarray inner gap} / \pgfkeysvalueof{/tikz/zoomboxarray columns} ) + \pgflinewidth
        }
        \pgfmathsetmacro\ypos{
            (\rowcount-1) * (\imageheight / \pgfkeysvalueof{/tikz/zoomboxarray rows} + \pgfkeysvalueof{/tikz/zoomboxarray inner gap} / \pgfkeysvalueof{/tikz/zoomboxarray rows} ) + 0.5*\pgflinewidth
        }
        \edef\dospy{\noexpand\spy [
            #1,
            zoombox paths/.append style={
                black and white pattern=\patternnumber
            },
            every spy on node/.append style={#1},
            x=\imagewidth,
            y=\imageheight
        ] on (#2) in node [anchor=north west] at ($(zoomboxes container.north west)+(\xpos pt,-\ypos pt)$);}
        \dospy
        \pgfmathtruncatemacro\pgfmathresult{ifthenelse(\columncount==\pgfkeysvalueof{/tikz/zoomboxarray columns},\rowcount+1,\rowcount)}
        \global\let\rowcount=\pgfmathresult
        \pgfmathtruncatemacro\pgfmathresult{ifthenelse(\columncount==\pgfkeysvalueof{/tikz/zoomboxarray columns},1,\columncount+1)}
        \global\let\columncount=\pgfmathresult
        \ifblackandwhitecycle
            \pgfmathtruncatemacro{\newpatternnumber}{\patternnumber+1}
            \global\edef\patternnumber{\newpatternnumber}
        \fi
    \end{scope}
}

\newcommand\zoomboxfusion[2][]{
    \begin{scope}[zoombox paths]
        \pgfmathsetmacro\xpos{
            (\columncount-1)*(\imagewidth / \pgfkeysvalueof{/tikz/zoomboxarray columns} + \pgfkeysvalueof{/tikz/zoomboxarray inner gap} / \pgfkeysvalueof{/tikz/zoomboxarray columns} ) + \pgflinewidth
        }
        \pgfmathsetmacro\ypos{
            (\rowcount-1) * (\imageheight / \pgfkeysvalueof{/tikz/zoomboxarray rows} + \pgfkeysvalueof{/tikz/zoomboxarray inner gap} / \pgfkeysvalueof{/tikz/zoomboxarray rows} ) + 0.5*\pgflinewidth
        }
        \edef\dospy{\noexpand\spy [
            #1,
            zoombox paths/.append style={
                black and white pattern=\patternnumber
            },
            every spy on node/.append style={#1},
            x=\imagewidth,
            y=\imageheight
        ] on (#2) in node [anchor=north west] at ($(zoomboxes container.north west)+(\xpos pt,-\ypos pt)$);}
        \dospy
        \pgfmathtruncatemacro\pgfmathresult{ifthenelse(\columncount==\pgfkeysvalueof{/tikz/zoomboxarray columns},\rowcount+1,\rowcount)}
        \global\let\rowcount=\pgfmathresult
        \pgfmathtruncatemacro\pgfmathresult{ifthenelse(\columncount==\pgfkeysvalueof{/tikz/zoomboxarray columns},1,\columncount+1)}
        \global\let\columncount=\pgfmathresult
        \ifblackandwhitecycle
            \pgfmathtruncatemacro{\newpatternnumber}{\patternnumber+1}
            \global\edef\patternnumber{\newpatternnumber}
        \fi

    \end{scope}
}

\newcommand\zoomboxXT[2][]{
    \begin{scope}[zoombox paths]
        \pgfmathsetmacro\xpos{
            (\columncount-1)*(\imagewidth / \pgfkeysvalueof{/tikz/zoomboxarray columns} + \pgfkeysvalueof{/tikz/zoomboxarray inner gap} / \pgfkeysvalueof{/tikz/zoomboxarray columns} ) + \pgflinewidth
        }
        \pgfmathsetmacro\ypos{
            (\rowcount-1) * (\imageheight / \pgfkeysvalueof{/tikz/zoomboxarray rows} + \pgfkeysvalueof{/tikz/zoomboxarray inner gap} / \pgfkeysvalueof{/tikz/zoomboxarray rows} ) + 0.5*\pgflinewidth
        }
        
        \edef\dospy{\noexpand\spy [
            #1,
            zoombox paths/.append style={
                black and white pattern=\patternnumber
            },
            every spy on node/.append style={#1},
            x=\imagewidth,
            y=\imageheight
        ] on (#2) in node [anchor=north west] at ($(zoomboxes container.north west)+(\xpos pt,-\ypos pt)$);}
        \dospy
        \pgfmathtruncatemacro\pgfmathresult{ifthenelse(\columncount==\pgfkeysvalueof{/tikz/zoomboxarray columns},\rowcount+1,\rowcount)}
        \global\let\rowcount=\pgfmathresult
        \pgfmathtruncatemacro\pgfmathresult{ifthenelse(\columncount==\pgfkeysvalueof{/tikz/zoomboxarray columns},1,\columncount+1)}
        \global\let\columncount=\pgfmathresult
        \ifblackandwhitecycle
            \pgfmathtruncatemacro{\newpatternnumber}{\patternnumber+1}
            \global\edef\patternnumber{\newpatternnumber}
        \fi
    \end{scope}
}

\definecolor{dnerf}{HTML}{FFC107}
\definecolor{tnvs}{HTML}{9948BF}
\definecolor{tnvb}{HTML}{4A1B41}
\definecolor{v4d}{HTML}{D81B60}

\definecolor{kp}{HTML}{1E88E5}
\definecolor{hp}{HTML}{C4C406}
\definecolor{fdgs}{HTML}{2A5D16}

\usepackage[capitalize]{cleveref}

\def\BibTeX{{\rm B\kern-.05em{\sc i\kern-.025em b}\kern-.08em
    T\kern-.1667em\lower.7ex\hbox{E}\kern-.125emX}}
\begin{document}

\title{WavePlanes: Compact Hex Planes for Dynamic Novel View Synthesis}

\author{Adrian Azzarelli, Nantheera Anantrasirichai, David R Bull\\
University of Bristol\\
United Kingdom\\
{\tt\small \{a.azzarelli, n.anantrasirichai, dave.bull\}@bristol.ac.uk}}

\maketitle

\begin{abstract}
Dynamic Novel View Synthesis (Dynamic NVS) enhances NVS technologies to model moving 3-D scenes. However, current methods are resource intensive and challenging to compress. To address this, we present WavePlanes, a fast and more compact hex plane representation, applicable to both Neural Radiance Fields and Gaussian Splatting methods. Rather than modeling many feature scales separately (as done previously), we use the inverse discrete wavelet transform to reconstruct features at varying scales. This leads to a more compact representation and allows us to explore wavelet-based compression schemes for further gains. The proposed compression scheme exploits the sparsity of wavelet coefficients, by applying hard thresholding to the wavelet planes and storing nonzero coefficients and their locations on each plane in a Hash Map. Compared to the state-of-the-art (SotA), WavePlanes is significantly smaller, less resource demanding and competitive in reconstruction quality. Compared to small SotA models, WavePlanes outperforms methods in both model size and quality of novel views.
\end{abstract}

\begin{IEEEkeywords}
Dynamic Novel View Synthesis, Neural Radiance Fields, Gaussian Splatting
\end{IEEEkeywords}

\section{Introduction}
\label{sec:intro}
\begin{figure*}[ht]
    \centering
    \begin{subfigure}[tb]{0.15\linewidth}
        \plotcroptrex{kplane_exp.jpg}{200MB}{K-Planes}{31.28}{0.980}
    \end{subfigure}
    \hspace{-2mm}
    \begin{subfigure}[tb]{0.15\linewidth}
        \plotcroptrex{waveplanes.jpg}{\underline{15MB}}{Ours-NeRF}{31.46}{0.979}
    \end{subfigure}
    \hspace{-2mm}
    \begin{subfigure}[tb]{0.15\linewidth}
        \plotcroptrex{4dgs.png}{33MB}{4D-GS}{34.23}{0.985}
    \end{subfigure}
    \hspace{-2mm}
    \begin{subfigure}[tb]{0.15\linewidth}
        \plotcroptrex{waveplanesgs.png}{27MB}{Ours-GS}{\underline{35.88}}{\underline{0.988}}
    \end{subfigure}
    \begin{subfigure}[tb]{0.15\linewidth}
        \plotcroptrex{gt.png}{X}{Ground Truth}{X}{X}
    \end{subfigure}
    \begin{subfigure}[tb]{0.20\linewidth}
        \begin{tikzpicture}[overlay, remember picture]
            \node[anchor=south west, black] at (-.0, -2.){\adjincludegraphics[width=0.95\linewidth] {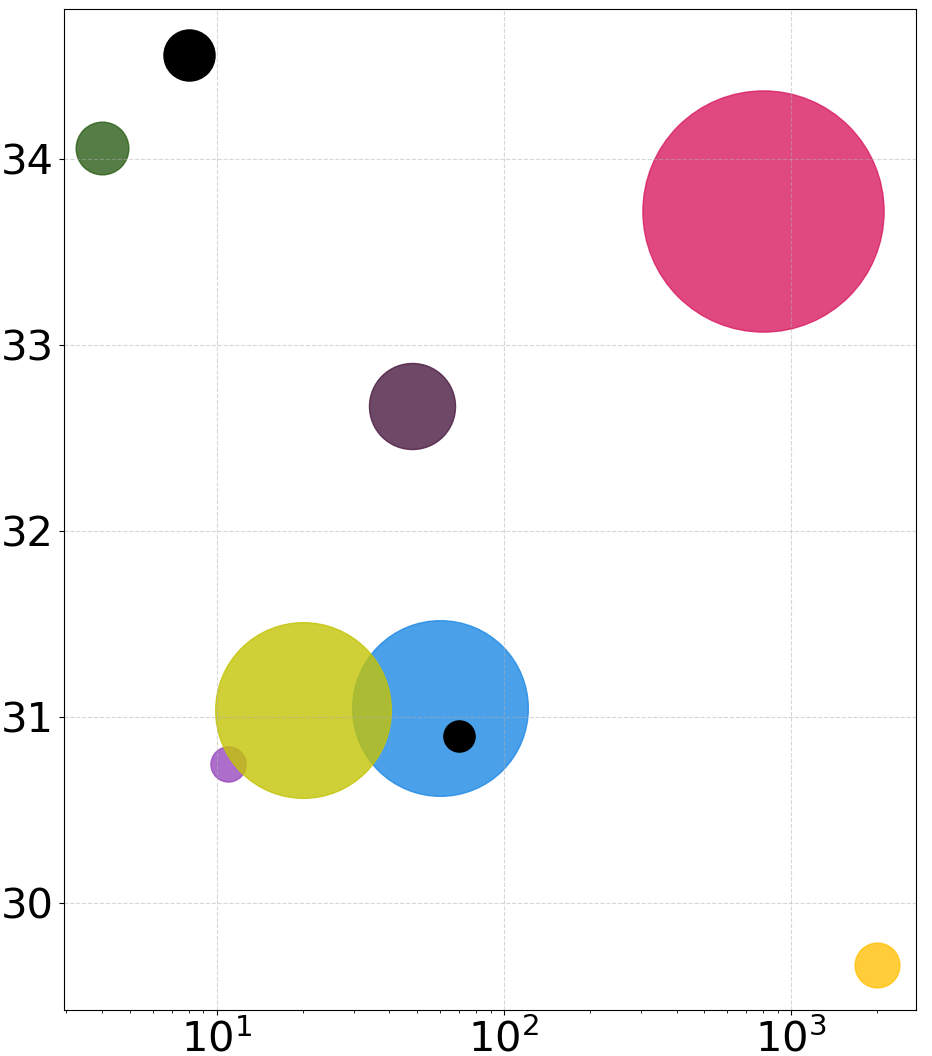}};
            \node[anchor=south west, rotate=90, black] at (.2, 1.1){\scriptsize PSNR};
            \node[anchor=south west, black] at (1.8, -2.3){\scriptsize Time (minutes)};
            
            \node[anchor=south west, color=dnerf] at (2.45, -1.5){\scriptsize D-NERF};
            \node[anchor=south west, color=v4d] at (2.6, 1.6){\scriptsize V4D};
            \node[anchor=south west, color=tnvs] at (.3, -1.3){\scriptsize TiNeuVox-s};
            \node[anchor=south west, color=tnvb] at (1., 0.6){\scriptsize TiNeuVox-B};
            \node[anchor=south west, color=kp] at (1.6, -.3){\scriptsize K-Planes};
            \node[anchor=south west, color=hp] at (.35, -.35){\scriptsize Hex-Planes};
            \node[anchor=south west, color=fdgs] at (0.55, 1.25){\scriptsize 4D-GS};
            \node[anchor=south west, color=black] at (0.85, 1.62){\scriptsize Ours-GS};
            \node[anchor=south west, color=black] at (1.85, -0.8){\scriptsize Ours-NeRF};
        \end{tikzpicture}
    \end{subfigure}
    \caption{{Left:} Visual and quantitative comparison of our method applied to K-Planes \cite{fridovich2023k} and 4D-GS \cite{wu20244dgs}. {Right:} Quantitative comparison of model size (point radius), training time (x-axis) and quality (y-axis) on the entire D-NeRF data set\cite{pumarola2021d}}
    \label{fig:frontpage}
\end{figure*}%
Novel view synthesis (NVS) has attracted significant attention with recent developments in 3-D scene generation typically employing Neural Radiance Field (NeRF) or Gaussian Splatting (GS) representations. Consequently, Dynamic NVS extends this by modeling an additional time component to support the generation of realistic object and scene animations. While static frameworks yield impressive results, dynamic NeRF/GS frameworks face limitations with computational complexity, scalability and compression.

In achieving dynamic NVS, plane-based representations have become a common way of explicitly decoding 4-D feature representations into low rank and compact 2-D components \cite{shao2023tensor4d, cao2023hexplane, huang2023sc}. This breakthrough was popularized by K-Planes \cite{fridovich2023k} for NeRF modeling, which led to 4D-GS \cite{wu20244dgs} the first dynamic GS solution. While faster in training than non-plane-based alternatives, these models tend to be computationally costly and/or challenging to compress, making tasks such as file-sharing and streaming problematic. 
For static NVS, wavelet-based tri-plane representations have been proposed to resolve these issues \cite{rho2023masked, xu2023wavenerf}. However, these methods do not translate to the case of dynamic NVS, where temporal features are interpreted differently to spatial features; hence they  cannot be compressed in the same manner.

In this paper, we propose a novel dynamic plane representation and compression method that uses wavelets as a base representation. This reduces computation during training and exploits the sparsity of the wavelet representation to compress the  model after training. Our pipeline decomposes a 4-D scene into a set of six 2-D grids and uses the 2-D inverse discrete wavelet transform (IDWT) to recover two sets feature planes from the 2-D wavelet representation (note \cite{shao2023tensor4d, fridovich2023k, cao2023hexplane} model four sets). This shifts the learning of the multi-scale representation to the IDWT, thus lowering GPU utilization and  enabling the modeling of complex data sets without the need of high performance computing resources.

Unlike prior approaches for compressing static wavelet representations, our method does not introduce any additional learnable parameters. Instead, we exploit the spatial and temporal sparsity, typical  in 4-D scenes, to compress the model.

In the above context, our main contributions are:
\begin{enumerate}[leftmargin=13pt, nolistsep]
\item The first dynamic hex plane representation that recovers features stored as 2-D wavelet coefficients using the IDWT. This approach (i) conditions the wavelet coefficients to be sparse - necessary for higher compression rates while retaining high quality, and (ii) is applicable to both NeRF and GS plane-based models.

\item A novel wavelet-based compression scheme that compresses NeRF and GS hex plane representations to SotA model size with no loss in quality.
\end{enumerate}

\section{Related Work}\label{sec: related works}
\paragraph*{NVS Background}
NVS pipelines require several components for robust 3-D reconstruction: (i) A data set containing extrinsic and intrinsic parameters; (ii) A volume sampler to model 3-D geometry; (iii) A method to learn per volume color and density features; and (iv) A method to render novel view using (ii) and (iii). Typically, (ii) and (iii) involve learning mechanisms.
Hence, the focus of most NVS methods is an improved, learnable, visual or geometric feature representations. This is challenged by quality, inference speed, computation and model size. For example, the key differences between 4D-GS and K-Planes are the methods for volumetric sampling and rendering. K-Planes uses the approach proposed by \cite{barron2022mip}, while 4D-GS relies on a point cloud representation to model geometry. Ultimately, 4D-GS produces better visual results and is faster, though it is less robust with geometry and prone to rendering artifacts \cite{yu2024mip}. 
In practice, both methods are challenging to compress and typically require a significant amount of RAM and/or GPU memory. Hence, we investigate a method for reducing model size and computation for both NeRF and GS use cases.

\paragraph*{Dynamic NVS} Originally, \cite{pumarola2021d} and \cite{li2022neural} proposed dynamic 3-D representations that leverage 3-D deformation fields to model per point/per volume positional change w.r.t time. Overcoming limitations with fixed scene topology, \cite{shao2023tensor4d, fridovich2023k, cao2023hexplane} proposed decomposing the 4-D feature space into bi-vector (2-D grid) representations. Key-frame representations alleviate similar problems \cite{song2023nerfplayer, attal2023hyperreel, icsik2023humanrf}, whereby methods focus on interpolating sequential static NeRF representations to make dynamic inference. Other NVS methods also exist to directly tackle temporally conditioned 3-D primitives, \cite{li2023dynibar, gan2023v4d}, though these methods have more nuanced limitations.

Recently, research into 4-D NVS has investigated dynamic GS representations \cite{wu20244dgs, yang2023deformable3dgs, huang2023sc}. Many works adopt NeRF-based 4-D plane representations for dynamic inference. For example, 4D-GS uses K-Planes to model point-based deformations relating to position, scale and rotation. Motivated by recent developments, we investigate a method that is applicable to both NeRF and 4D-GS.

\paragraph*{Wavelet NeRFs} 
Wavelet-based modeling of static scenes has been shown to improve the compactness of existing representations \cite{saragadam2023wire, rho2023masked, xu2023wavenerf}.  
For example, the masked wavelet transform (MWT) method \cite{rho2023masked} introduces a tri-plane decomposed wavelet-based representation that enhances compression via a learned binary mask for each plane to exploit the sparsity of the representation. However, it is challenging to apply this method to dynamic NeRFs as it: (i) Requires a high level wavelet decomposition even for static scenes, (ii) Requires a learnable binary mask for each plane at varying scales/resolutions and (iii) introduces significant complexity overheads.

Instead, we introduce the first wavelet-based representation for modeling dynamic NVS. Unlike MWT, where compression relies on the training algorithm's ability to decode the binary mask, our approach does not require learning additional parameters - our compression scheme only needs to be executed once. Furthermore, while other dynamic plane representations use four sets of feature scales \cite{fridovich2023k, cao2023hexplane}, we only produce two sets of features (discussed in \cref{sec: method}) making our method more compact during training.


\section{WavePlanes}\label{sec: method}
\subsection{Hex Plane Decomposition}\label{sec: hex plane background}
We describe below how 4-D data is decomposed into the set of possible feature-based bi-vectors, $\mathbf{P^r_c}$ where $c \in C = \{xy, xz, yz, xt, yt, zt\}$ and $r$ is the scale, for NVS tasks. 

A volumetric sampler samples a volume's position $\mathbf{q} = [x, y, z, t]$ in world space. The planes $\mathbf{P^r_c}$ are orthogonal and axis-aligned to the world space. The point is then projected onto each plane using $\pi_c(\mathbf{q})$ and bi-linearly interpolated w.r.t the 2-D grid representations using $f(\mathbf{q})_c^r =  \psi(\mathbf{P}_c^r, \pi_c(\mathbf{q}))$. This returns a feature, $f(\mathbf{q})_c^r$, for each bi-vector and scale. Subsequently, all (six) features for a given point are fused, e.g. K-Planes select an element-wise product. The fused feature is then linearly decoded into the desired parameter. For NeRF, this is volume color and density \cite{fridovich2023k, cao2023hexplane, shao2023tensor4d}. For GS it may be volume position, scale and rotation deformation \cite{wu20244dgs}.

\subsection{Proposed Method Overview}
\begin{figure*}[tb]
  \centering
   \includegraphics[width=\linewidth]{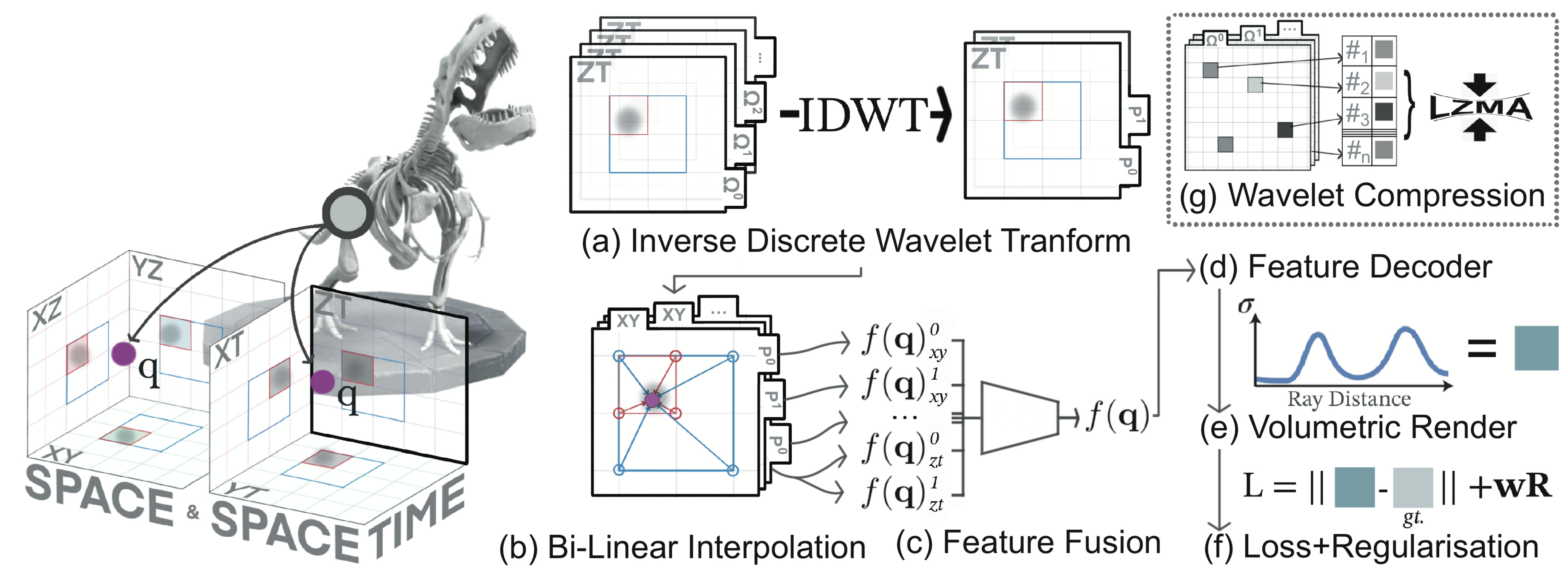}
   \caption{{Model Pipeline:} (a) N-level wavelet planes are transformed using the IDWT into feature planes. (b) 4-D samples are projected onto each plane and bi-linearly interpolated over multiple scales. (c) Volumetric features are recovered by fusing the space-only and space-time features. (d) Features are linearly decoded into color and density values. (e) A 2-D pixel is rendered from the 3-D volume using the NeRF volumetric rendering function. (f) A loss and weighted regularization is used for training.  (g) \textit{After training} we compress nonzero wavelet coefficients}
   \label{fig:architecture}
\end{figure*}
The hex plane model for WavePlanes is shown in \cref{fig:architecture}. The representation is stored as a set of 2-D wavelet coefficients, $\mathbf{\Omega}^s_c$ where $s \in [0, 1, ..., N]$ denotes the wavelet level and $c$ was previously defined. $\mathbf{\Omega}^s_c$ is encoded using the IDWT into a fine feature plane, $\mathbf{P}_c^r$ where $r=0$, and a coarse feature plane where $r=1$, as shown in \cref{fig:architecture} (a). This is accomplished by performing the IDWT twice with the full set (all $N$-levels) and a reduced set ($N-1$ levels) of wavelet coefficients, respectively. Typically, four scales of feature planes are modeled \cite{fridovich2023k, cao2023hexplane}, however we shift the modeling of multi-scale features to the wavelet representation, which allows us to reduce GPU utilization by only modeling two scales of feature planes. Steps (b-f) in \cref{fig:architecture} were discussed in \cref{sec: hex plane background}. 

To compress our wavelet representation, we first filter out zero-valued wavelet coefficients and store the remaining nonzero coefficients and their locations in $\mathbf{\Omega}^s_c$ as a Hash Map, after training. We then apply lossless compression to reduce to model size, as shown in \cref{fig:architecture} (g).

\subsection{Feature Plane Reconstruction}
The 2-D wavelet coefficient planes are separated into low frequency coefficients, representing a low resolution approximation of the feature planes $\mathbf{P}^0_c$, and high frequency coefficients, for modeling fine details at various frequency bands. We refer to the low frequency wavelet planes as $\Omega_c^{[s=0]} \in \mathcal{R}^{B \times H/2^N \times W/2^N}$, where $B=64$ is the feature length in this paper, $(H, W)$ is the desired height and width of the reconstructed feature plane and $2^N$ is a downsampling factor involved in wavelet decomposition. 
The high frequency wavelet coefficients are defined as $\Omega_c^{[s>0]}  \in$\linebreak$\mathcal{R}^{B\times 3 \times H/s' \times W/s'}$, where 3 indicates the number of filters\footnote{Horizontal (LH), vertical (HL) and diagonal (HH) filters.} and $s' = 2^{N-s}$ allows for downsampling where the lowest and highest frequency planes will have size $(H/2^N, W/2^N)$ and $(H/2,W/2)$ respectively.
Hence, for a wavelet decomposition level of 2, $N=2$ and $s \in [0,1,2]$. 

To reconstruct the feature planes we use \eqref{eq: idwt function}. To avoid vanishing gradients for high-frequency coefficients, we apply a frequency-based scaling factor to each plane, denoted by $\mathbf{k} \odot \mathbf{\Omega}_c$ as suggested in \cite{rho2023masked}, where we select $\mathbf{k} = [1, \frac{2}{5}, \frac{1}{5}]$.
\begin{equation}\label{eq: idwt function}
    P_c^{s}(\Omega_c|s)
    \begin{cases}
        \text{IDWT}(\mathbf{k} \odot \Omega_c^{[0:s]}),& \text{if } c\in [xy,xz,yz],\\
        \text{IDWT}(\mathbf{k} \odot \Omega_c^{[0:s]} )+1,              & \text{otherwise}.
    \end{cases}
\end{equation}

Here the condition biases the space-time features towards 1. As space-time mostly consists of null volumes, introducing the $+1$ term is a simple way of ensuring robust initialization and regularization while preserving wavelet sparsity. As 3-D space conditioned on time mostly consists of static empty volumes, allowing the space-only features to tend to 0 (rather than 1) maximizes the sparsity of space-only wavelets.

\subsection{Compressing Wavelets with Hash Maps}\label{sec: model.compressing}
A benefit of using wavelets is the ability to compress representations by exploiting sparsity. All  planes exhibit this property due to the condition proposed in \eqref{eq: idwt function}.
To compress the representation, 0-valued wavelet coefficients are filtered out using hard thresholding. The nonzero coefficients and their individual locations on each wavelet plane, $\mathbf{\Omega}^s_c$, are then stored in a Hash Map. Finally, lossless compression is applied to further reduce the final size of the model. Conversely, to load the model from the compressed representations we initialize the wavelet planes with zero-values and then restore the nonzero coefficients to their original locations.

\paragraph*{Why not compress the feature planes?} While there are certainly fewer feature planes than wavelet coefficient planes, the majority of nonzero coefficients are concentrated at the lowest resolution (low frequency) wavelet planes, which are $2^N$ smaller than the highest resolution feature planes. Also, the highest resolution wavelet planes are $1/4$ the size of the feature planes, and are the sparsest set of high frequency wavelet coefficients.

\paragraph*{Why hard thresholding?} Modifying nonzero coefficients, for instance with soft thresholding, produces visually incomprehensible renders as the linear decoder is sensitive to minor changes in the feature space. Nonetheless, filtering out near-zero coefficients can reduce noise and promote sparsity.

\paragraph*{Why Hash Maps?} These enable storing and loading nonzero wavelet coefficients efficiently, and support  further compression using lossless schemes. Here, we selected the Lempel–Ziv–Markov chain (LZMA) scheme as it provides the best compression (see supplementary experiments).

\subsection{Optimization}\label{sec:optimization}
\paragraph*{Regularization} For all experiments we use three regularizers: total variation (TV), spatial smoothness in time (SST) and time sparsity (TS). TV, in \eqref{eq: tv regularizer}, is well known for achieving smoother signal representations \cite{estrela2016total}.
\begin{equation}\label{eq: tv regularizer}
    \mathcal{L}_{TV}(\mathbf{P}) = \frac{1}{|C|n^2}\sum_{c,i,j} (|| \mathbf{P}^{i,j}_c - \mathbf{P}^{i-1,j}_c ||^2_2 + || \mathbf{P}^{i,j}_c - \mathbf{P}^{i,j-1}_c ||^2_2)
\end{equation}

SST, in \eqref{eq: sst regularizer}, represents the 1-D Laplacian approximation of the second derivatives of the spatial components for each time-plane.  
Rather than smoothing along the time-axis as proposed in \cite{fridovich2023k}, smoothing along the spatial axis' preserves details and avoids over smoothing in the presence of fast motion, leading to better perceptual results. 
\begin{equation}\label{eq: sst regularizer}
    \mathcal{L}_{SST} = \frac{1}{|C|n^2}\sum_{c,i,t} || \mathbf{P}^{i-1,t}_c - 2\mathbf{P}^{i,t}_c + \mathbf{P}^{i+1,t}_c ||^2_2
\end{equation}

TS, proposed in \eqref{eq: st regularisation}, uses the $\ell_1$-norm to promote sparsity, maximizing the number of zero-valued space-time wavelet coefficients. 
This is useful for compression as it directly maximizes the sparsity of $\mathbf{\Omega}^s_{c_t}$.
\begin{equation}\label{eq: st regularisation}
    \mathcal{L}_{TS} = \sum_{c_t \in C_t} ||\mathbf{\Omega}_{c_t}||_1
\end{equation}

\paragraph*{Feature decoder with learned color basis} For NeRFs, to decode volumetric features into color and density values, a per scene learned color basis \cite{wizadwongsa2021nex} was proposed, where features are treated as coefficients for a linear decoder. We utilize the same approach for the NeRF variant of our model. This maps the viewing direction of a camera ray $\mathbf{d}$ to red, green and blue basis vectors, $b_i(\mathbf{d})$ for $i \in \{R,G,B\}$. 
An additional density basis is learned and left independent of viewing direction. Subsequently, color and density values are recovered by applying the dot product between basis vectors $b_i(\mathbf{d})$ and the fused feature $f(\mathbf{q})$.

\paragraph*{Volumetric Sampling Network} WavePlanes is applicable to both NeRF and GS style volumetric samplers. For the NeRF version, we adapt the Mip-NeRF 360 \cite{barron2022mip} approach and train a lower resolution version of WavePlanes to learn the approximate distribution of volumes per ray in time. For GS, we learn the distribution of volumes by modeling each volume as a 3-D Gaussian blob, with temporally conditioned scale, rotation and position.

\paragraph*{Importance Sampling} For NeRFs, importance sampling (IST) (proposed by \cite{li2022neural}) is typically used to optimize ray selection while training on the DyNeRF data set \cite{li2022neural}. Normally, images are downsampled by a factor of 2 and rays with higher temporal importance are attributed to 4-D volumes that are assumed to be dynamic. However, due to computational constraints we were unable to pre-generate the full IST weights in our experiments. Instead, we downsampled images by a factor of 8. For fair comparison, we ran K-Planes-Compact on the same set-up. This is a reduced version of K-Planes that consists of only two scales (rather than four) with the same resolution as the two feature planes produces by the IDWT; shown in \cref{fig:architecture} (a).

\paragraph*{Training Speed} To implement the wavelet functionality we use the \textit{pytorch\_wavelets} library \cite{cotter2020uses}. 
To boost training speed we cache (on the GPU) the resulting feature planes, $\mathbf{P}_c^s$, at the start of every epoch to avoid re-processing the IDWT during regularization. 
This significantly reduces training time but slightly increases memory consumption, shown in \cref{tab:gpu comparison with kplanes}.
\begin{table}[tb]
    \centering
    \caption{{Computational comparisons} of our NeRF model with and without caching every epoch. Note: It was not possible to execute K-Planes for the DyNeRF scenes} 
    \begin{tabular}{c c c c}
        \toprule
        Method & GPU Mem. $\downarrow$ & GPU Util. $\downarrow$ &  Time $\downarrow$  \\  \midrule
        \multicolumn{4}{c}{Synthetic D-NeRF assets \cite{pumarola2021d}}   \\ \midrule
        K-Planes \cite{fridovich2023k} & 7.0 GB & 86\% & 59 mins\\
        Ours & 7.4 GB & 70\% & 120 mins \\ 
        Ours + Caching & 7.7 GB & 79\%  & 72 mins \\ \midrule
        \multicolumn{4}{c}{Real DyNeRF scenes \cite{li2022neural}} \\ \midrule
        Ours & 9.1 GB & 82\% & 690 mins \\
        Ours + Caching  & 9.7 GB & 89\% & 510 mins \\
        \bottomrule
    \end{tabular}
    
    \label{tab:gpu comparison with kplanes}
\end{table}

\section{Results and discussion}\label{sec: results}
\begin{figure*}[tb]
    \centering
    \def\arraystretch{1}
    \begin{tabular}{@{}c@{}c@{}c@{}c@{}c@{}c@{}}
        \videozoomed{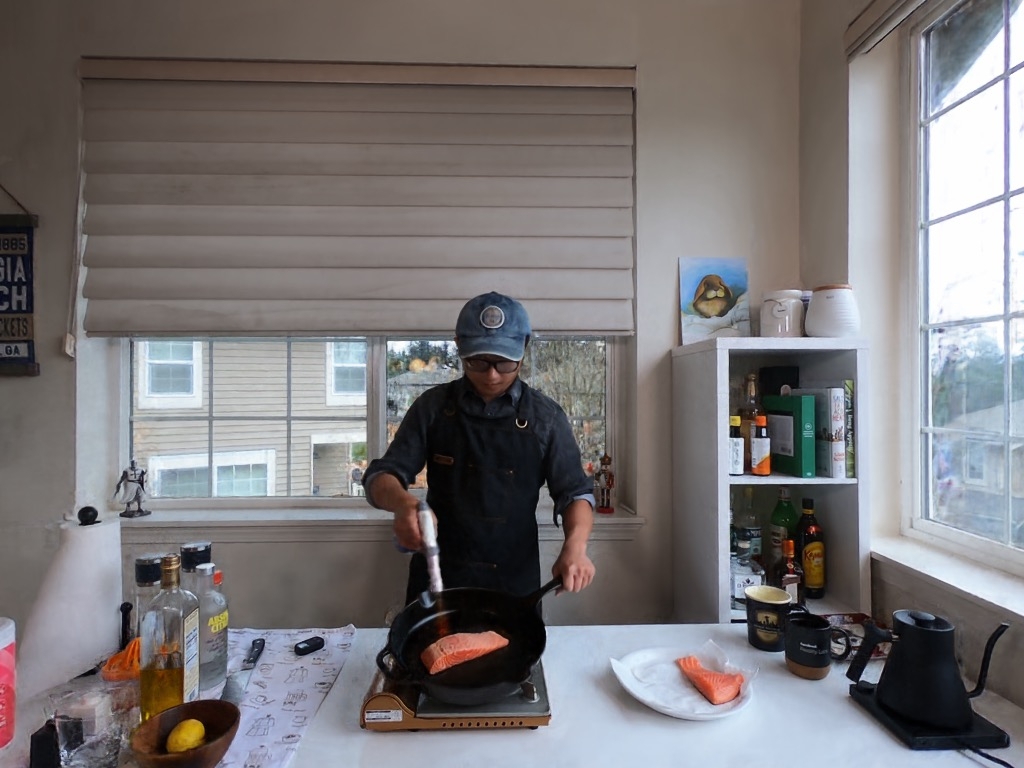}{200}{2} & 
        \videozoomed{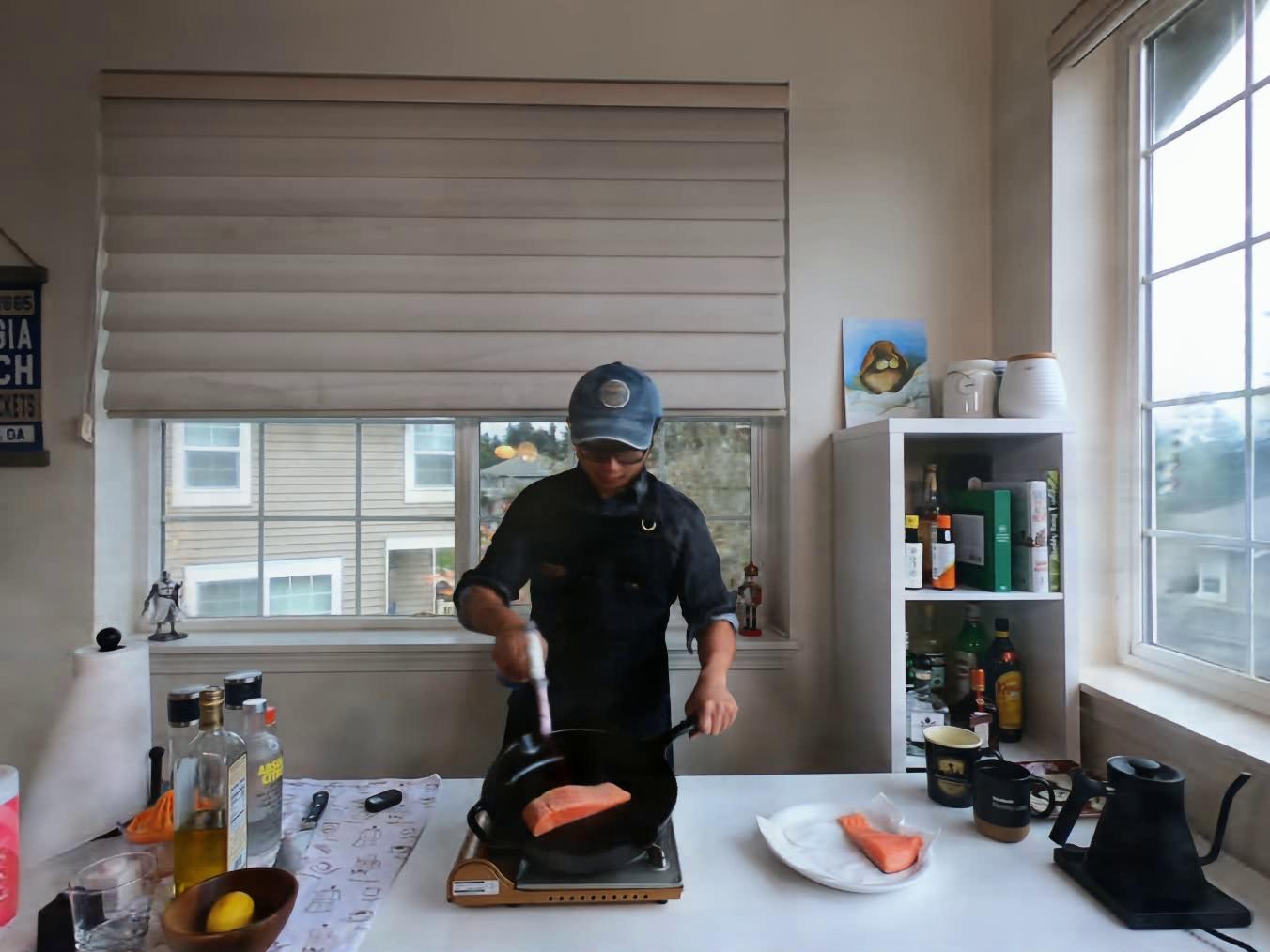}{250}{2}&
        \videozoomed{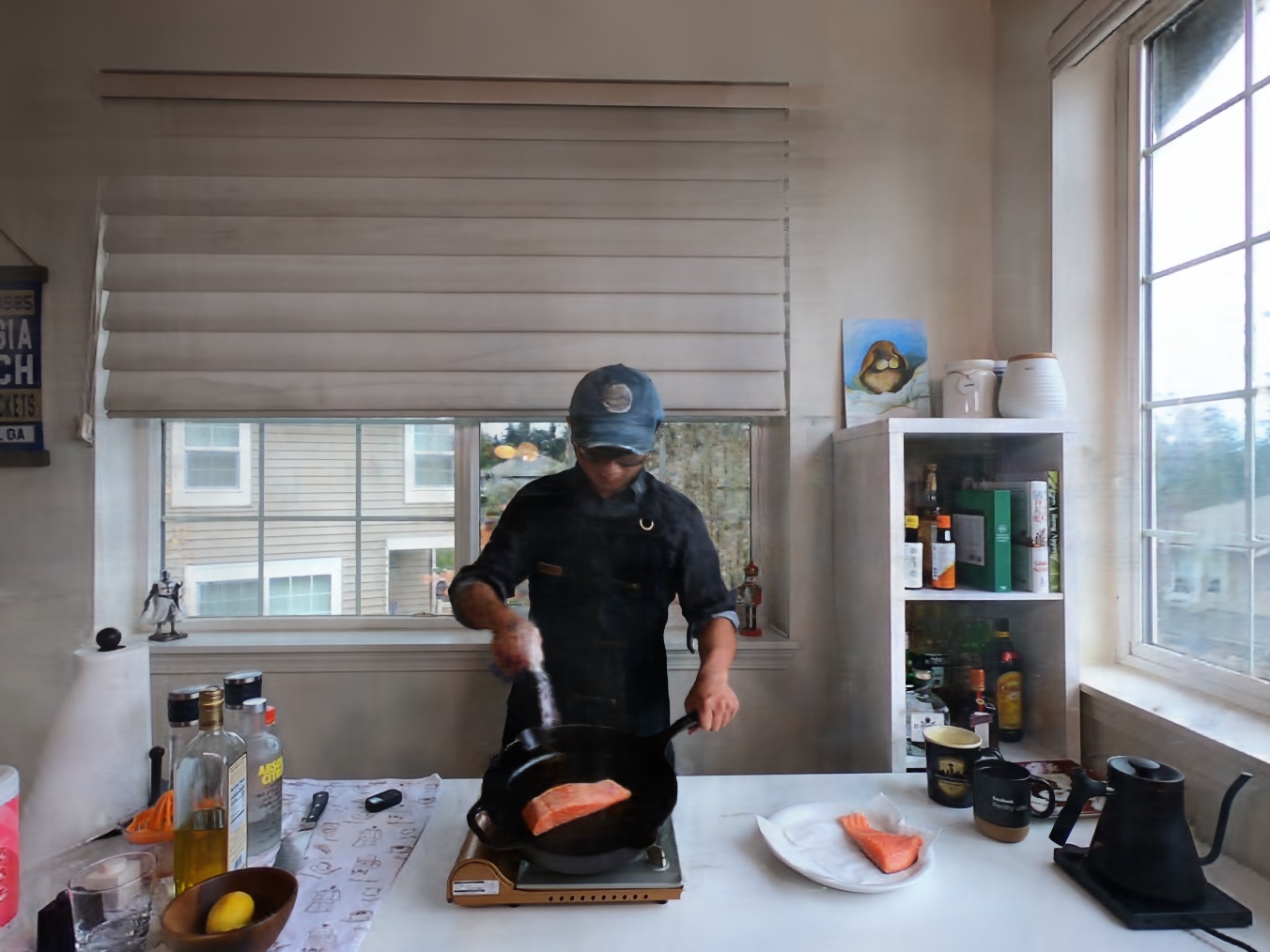}{57}{8} &
        \videozoomed{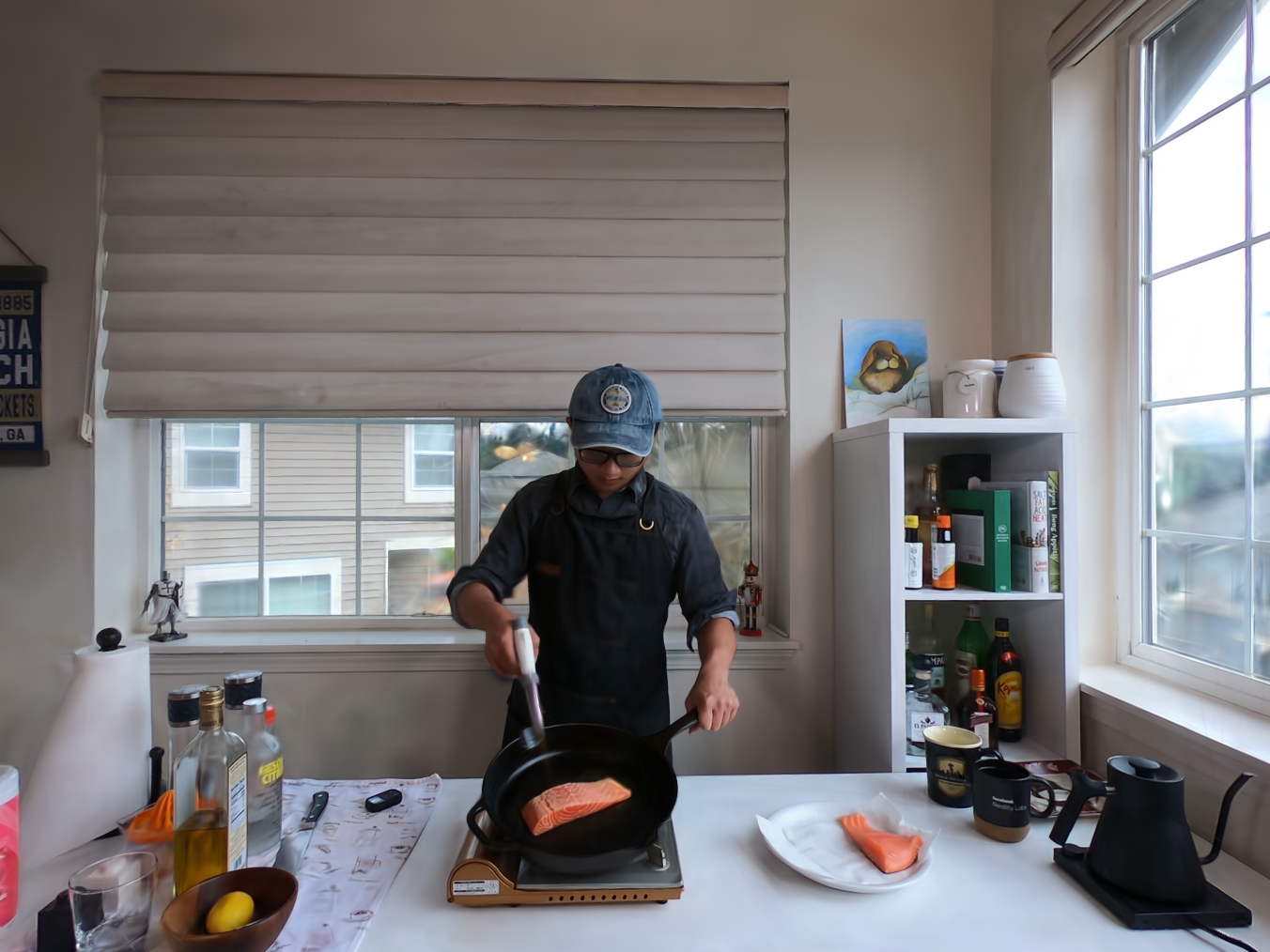}{36}{8} &
        \videozoomed{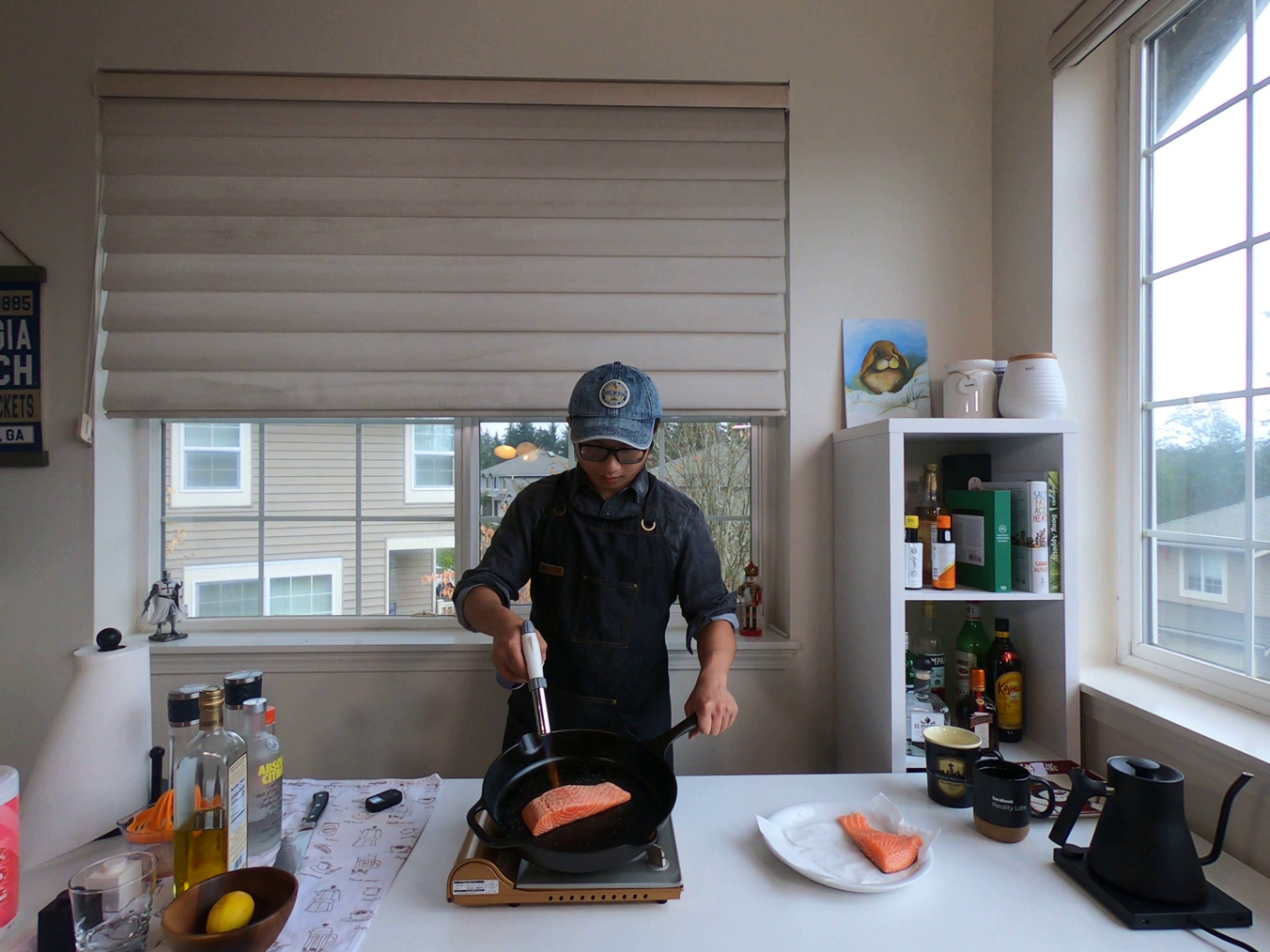}{X}{1}
        \\ [-12mm]
        \multicolumn{1}{c}{\small HexPlanes} &
        \multicolumn{1}{c}{\small K-Planes} &
        \multicolumn{1}{c}{\small Ours-NeRF} &
        \multicolumn{1}{c}{\small Ours-GS} &
        \multicolumn{1}{c}{\small Ground Truth}
    \end{tabular}
    \caption{{Qualitative real video results} on the Cooked Salmon DyNeRF scene \cite{li2022neural}. $2\times$ and $8\times$ indicates the downsampling factor used for the IST weights. Due to limited RAM we use $8\times$ downsampling}
    \label{fig: visual dynerf results}
    \vspace{3pt}
\end{figure*}
As our approach builds a neural representation rather than an NVS pipeline, our method can be applied to any plane-based NVS model. We demonstrate this by replacing the hex plane representation in K-Planes and 4D-GS with WavePlanes. These are labeled Ours-NeRF and Ours-GS, respectively. We compared WavePlanes with existing models, including D-NeRF \cite{pumarola2021d}, TiNeuVox \cite{fang2022fast}, HexPlanes \cite{cao2023hexplane}, K-Planes \cite{fridovich2023k}, V4D \cite{gan2023v4d} and 4D-GS \cite{wu20244dgs}.
In all these cases, we report the PSNR values cited in the corresponding original publications. For further comparison K-Planes-Compact was compressed using the proposed scheme by first turning the features into wavelet coefficients.

\begin{table}[tb]
    \centering
    \caption{Results on \textbf{synthetic} dynamic D-NeRF assets \cite{pumarola2021d}}
    \begin{tabular}{c c c c}
    \toprule
        Method & PSNR $\uparrow$ & SSIM $\uparrow$ & Size $\downarrow$  \\ \midrule
         \multicolumn{4}{c}{Large models} \\ \midrule
        HexPlanes \cite{cao2023hexplane} & 31.04 & N/A  &   200MB \\
        K-Planes \cite{fridovich2023k} & 31.05 & 0.97 &  200MB \\
        V4D \cite{gan2023v4d} & 33.72 & 0.97 & 377MB \\ \hline
        \multicolumn{4}{c}{Small models} \\ \hline
        4D-GS \cite{fridovich2023k} & 34.06 & 0.98 &  18MB \\
        D-NeRF \cite{pumarola2021d} & 29.67  &  0.95 &  13MB \\
        TiNeuVox \cite{fang2022fast} & 30.75 &  0.96 &  8MB \\ 
        Ours-NeRF & 30.90 &  0.97 & 6.3MB \\
        Ours-GS & 34.56 &  0.98 & 16.9MB \\
        \bottomrule
    \end{tabular}
    
    \label{tab: main results syn}
\end{table}

\begin{table}[tb]
    \centering
    \caption{Results on \textbf{real} dynamic DyNeRF scene \cite{li2022neural}}
    \begin{tabular}{c c c c c }
    \toprule
        Method &  PSNR $\uparrow$  & SSIM $\uparrow$  & Size $\downarrow$  \\ \midrule
        \multicolumn{4}{c}{Large models} \\ \midrule
        HexPlanes \cite{cao2023hexplane} & 31.65  & N/A  &  200MB\\
        K-Planes \cite{fridovich2023k} & 31.30  & 0.963  & 250MB \\ \midrule
        
        \multicolumn{4}{c}{Small models} \\ \midrule
        4D-GS \cite{wu20244dgs} & 31.91  & 0.947  & 90MB \\
        K-Planes-Compact & 28.83  & 0.915  & 71MB \\
        Ours-NeRF & 30.32  & 0.922  & 57MB \\
        Ours-GS & 31.59  & 0.940  & 36MB \\ \bottomrule   
    \end{tabular}
     
    \label{tab: main results real}
\end{table}

\paragraph*{Synthetic monocular scenes}
The D-NeRF data set \cite{pumarola2021d} is used for low and high complexity synthetic object animations. These were captured using monocular ``teleporting'' cameras rotated at 360$^{\circ}$ and contain 50-200 training frames. We tune the model per scene as some scenes perform better at lower resolution. The results in \cref{tab: main results syn} show Ours-NeRF achieves the smallest model size while outperforming other small NeRFs. This is evidenced by the visual comparisons in \cref{fig: visual dnerf lego} and \ref{fig: more_dnerf_zoomed}. It can be seen that the WavePlanes method improves the quality of details and produces fewer artifacts with significantly smaller model size. Moreover, when applied to GS, WavePlanes demonstrates SotA performance on asset generation while requiring only a small model size.

\paragraph*{Real multi-view scenes} The DyNeRF data set \cite{li2022neural} is used to test our model's ability to perform single-view synthesis on real dynamic scenes with 19 or 20 cameras, lasting 10 seconds at 30FPS. Quantitative results in \cref{tab: main results real} indicate competitive performance for Ours-NeRF, considering the reduced quality of the IST weight pre-generation. As shown in \cref{fig: visual dynerf results}, Ours-NeRF performs NVS well and shows sharper edges and less noise on a number of objects in the scene, e.g. the pattern on the uncooked salmon is clearer whilst those of K-Planes are smoothed out. Ours-GS also produces the smallest model size while retaining good rendering quality. We also show, in \cref{tab: main results real} and \cref{fig: 2kp} that our method outperforms K-Planes-Compact in quality and size. This demonstrates the importance of wavelets as a base representation for reducing computation during training.

\begin{figure}[tb]
  \centering
  \plotlego{Ground Truth}{TiNeuVox}{D-NeRF}{Ours}
  \caption{{Zoomed visual comparisons} of the Lego scene \cite{pumarola2021d} with PSNR / SSIM / model size}
  \label{fig: visual dnerf lego}
\end{figure}

\begin{figure}
 \centering
 \plotmorezoomed{0}
 \caption{Zoomed visual comparisons of small NeRFs on fast motion}
 \label{fig: more_dnerf_zoomed}
\end{figure}

\begin{figure}
 \centering
 \plotkp{0}
 \caption{Comparing render and error map of WavePlanes and K-Planes-Compact; both model 2 scaled feature representations}
 \label{fig: 2kp}
\end{figure}

\paragraph*{Ablation Study}
Results of an ablation is provided in \cref{tab: ablation}. Further details and results are provided in the supplemental material.
\begin{table}[tb]
    \centering
    \caption{Ablations on the DyNeRF scenes \cite{li2022neural}} 
    \begin{tabular}{c c c c c }
    \toprule
        Method &  PSNR $\uparrow$  & SSIM $\uparrow$  & Size $\downarrow$  \\ \midrule
        K-Planes-Compact & 28.83 & 0.915 & 200MB \\
        w/WavePlanes & 30.89 & 0.926 & 140MB \\ 
        w/WavePlanes + Compression & 30.89 & 0.926 & 58MB  \\ \midrule 
        4DGS & 31.91  & 0.940 &  90MB \\
        w/WavePlanes & 31.59  & 0.940  & 39MB \\ 
        w/WavePlanes + Compression & 31.59  & 0.940  & 36MB  \\ 
        \hline 
    \end{tabular}
    \label{tab: ablation}
\end{table}

\paragraph*{NeRF vs GS}
The quality of Ours-GS is significantly better than that of Ours-NeRF and for real scenes the final models size is smaller. However, the per scene results (see the Supplementary Materials) indicate edge cases where Ours-NeRF produces a significantly smaller model with a lower performance gap (e.g. the Lego and Balls scenes). This occurs when there are fewer images or the level of detail is lower in the training set and implies that Ours-NeRF is more robust to sparse datasets.

\paragraph*{Limitations}
As shown in \cref{tab: ablation}, our compression method is less suited to GS representations than NeRF representations. This is because the GS representation mainly consists of a point cloud representation that cannot be compressed with Hash Maps. Furthermore, for both NeRF and GS approaches, the compression method is limited by the density of objects and motions in scene. This explains the difference in performance gains between the denser DyNeRF scenes and sparse D-NeRF assets.

\section{Conclusion}
We introduce WavePlanes, the first wavelet-based representation for dynamic NVS applications. We demonstrate that WavePlanes is applicable to both NeRF and GS pipelines and is capable of synthesizing diverse scenes with low computational cost, competitive quality and pleasing visual results. By exploiting the characteristics of the wavelet representation, our novel compression scheme enables the production of small dynamic NVS models without loss in quality and without the need for additional training or parameters. The results demonstrate that using WavePlanes as a base feature representation can enhance various qualities of existing NeRF and GS pipelines. Applied to GS, our method is especially performant on dynamic asset generation tasks. Whereas for NeRF, our method is more robust to low detail or sparse data sets and can reduce model size up to $\times 4$ smaller than the next largest dynamic NeRF, while producing better visual and metric results.

\bibliographystyle{IEEEbib}
\bibliography{icme2025references}

\end{document}